\documentclass{article}

   \PassOptionsToPackage{numbers, compress}{natbib}

\usepackage[preprint]{neurips_2026}

\usepackage[utf8]{inputenc} 
\usepackage[T1]{fontenc}    
\usepackage{hyperref}       
\usepackage{url}            
\usepackage{booktabs}       
\usepackage{tabularx}       
\usepackage{amsfonts}       
\usepackage{nicefrac}       
\usepackage{microtype}      
\usepackage[table]{xcolor}
\usepackage{threeparttable}
\usepackage{multirow}
\usepackage{amsmath}        
\usepackage{algorithm}      
\usepackage{algorithmic}    
\usepackage{graphicx}
\usepackage{caption}
\usepackage{wrapfig}
\usepackage{amssymb}

\title{\raisebox{-0.35em}[0pt][0pt]{
\includegraphics[height=1.5em]{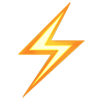}%
}Realtime-VLA \textcolor{orange!95!black}{FLASH}: Speculative Inference Framework for Diffusion-based VLAs}
\author{%
  {\bfseries Jiahui Niu$^{1,2}$\thanks{This work was done during the internship at Dexmal.}\quad Kefan Gu$^{3,4}$\quad Yucheng Zhao$^{4}$\thanks{Project lead.} \quad Shengwen Liang$^{1}$\thanks{Corresponding author.}\quad} \\[-0ex]
  {\bfseries Tiancai Wang$^{4}$\footnotemark[3]\quad Xing Hu$^{1}$\quad Ying Wang$^{1}$ \quad Huawei Li$^{1}$}
  \vspace{1mm} \\
  $^1$ State Key Lab of Processors, Institute of Computing Technology, CAS \\
  $^2$ University of Chinese Academy of Sciences \\
  $^3$ Nanjing University 
  $^4$ Dexmal \\
  \textit{Page:} \textcolor{orange!95!black}{\url{https://dexmal.github.io/realtime-vla-flash}}
}

\begin{document}

\maketitle

\vspace{-2.0em}
\begin{abstract}
\vspace{-1.0em}
Diffusion-based vision-language-action models (dVLAs) are promising for embodied intelligence but are fundamentally limited in real-time deployment by the high latency of full inference.
We propose \textbf{Realtime-VLA FLASH}, a speculative inference framework that eliminates most full inference calls during replanning by introducing a lightweight draft model with parallel verification via the main model's Action Expert and a phase-aware fallback mechanism that reverts to the full inference pipeline when needed.
This design enables low-latency, high-frequency replanning without sacrificing reliability.
Experiments show that on LIBERO, FLASH largely preserves task performance by replacing many 58.0\,ms full-inference rounds with speculative rounds as fast as \textbf{7.8\,ms}, lowering task-level average inference latency to 19.1\,ms (\textbf{3.04$\times$} speedup). We additionally demonstrate effectiveness on real-world conveyor-belt sorting, highlighting its practical impact for latency-critical embodied tasks.
\end{abstract}

\begin{figure}[h!]
  \vspace{-1em}
  \centering
  \includegraphics[width=0.75\linewidth]{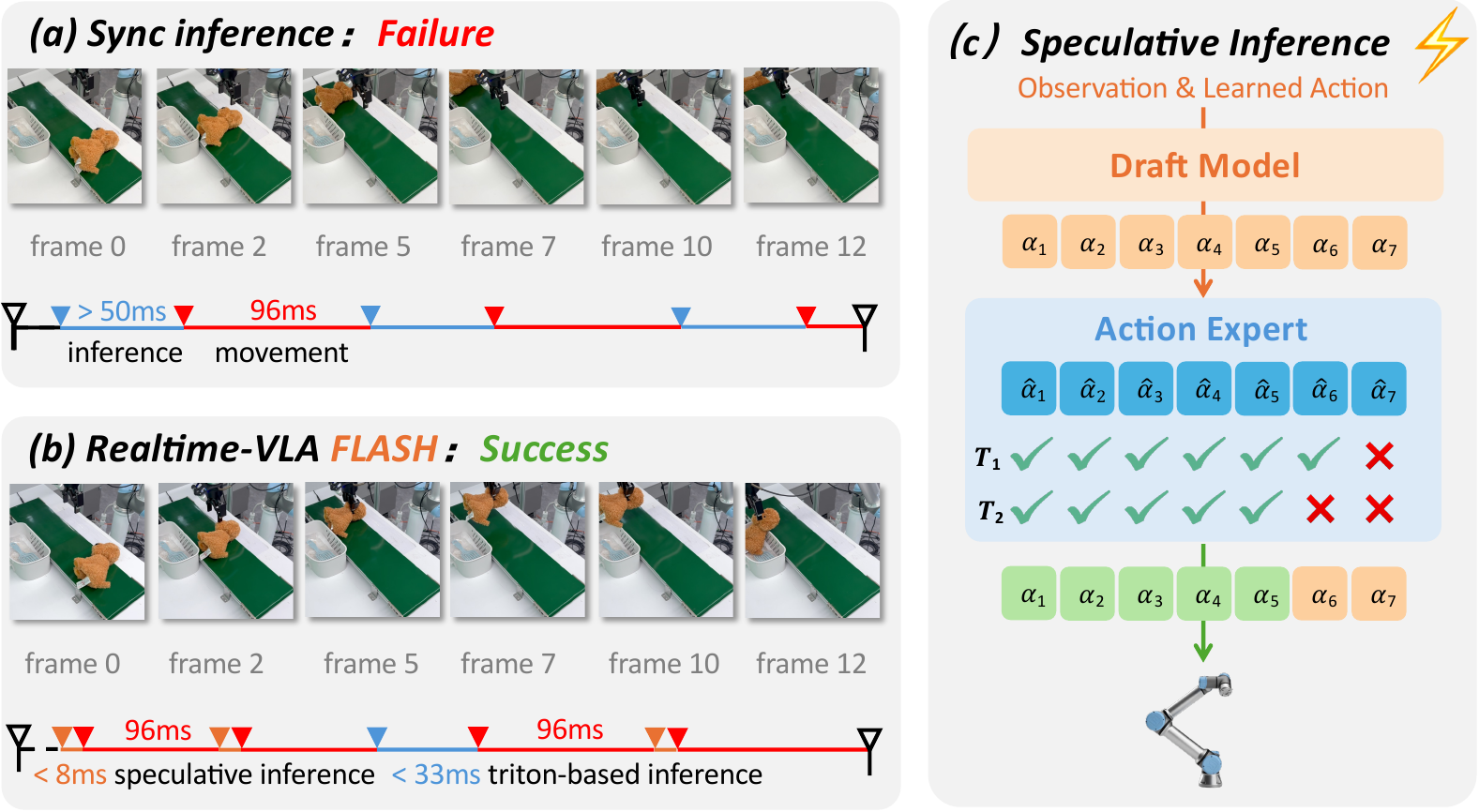}
  \caption{
  \textbf{Overview of Realtime-VLA FLASH.}
  (a) Standard synchronous dVLA inference runs the full pipeline at each replanning round, causing stale action updates and failure in latency-critical manipulation.
  (b) Realtime-VLA FLASH accelerates replanning, allowing the robot to react in time and complete the grasp.
  (c) FLASH introduces speculative inference for dVLAs: a lightweight draft model proposes a continuous action chunk, and the main model's Action Expert reconstructs action endpoints from selected flow-matching timesteps in parallel to check consistency with the draft. The robot executes the longest consistent prefix and falls back to the full path when verification fails.
}
  \label{fig:overview}
\end{figure}

\section{Introduction}
\label{sec:introduction}

Diffusion-based vision-language-action models (dVLAs) \cite{black2024pi_0, intelligence2025pi05visionlanguageactionmodelopenworld, gr00tn1_2025,gear2025gr00tn16} have recently emerged as an important paradigm for embodied intelligence. Their diffusion-based continuous action generation captures multimodal action distributions, making them well suited to complex manipulation tasks.

However, deploying dVLAs on real robotic systems remains challenging. In practice, robot control typically runs at a much higher frequency than model inference, as shown in Figure~\ref{fig:overview}, and existing systems therefore rely on open-loop action chunking \cite{zhao2023learning} to bridge this mismatch. For example, $\pi_0$ \cite{black2024pi_0} predicts a chunk of future actions and replans only after several control steps have already been executed. While this design reduces the frequency of replanning, each replanning round still requires the same expensive full inference pipeline. As a result, end-to-end inference latency remains the main bottleneck, limiting the applicability of dVLAs to reactive, latency-sensitive tasks.

Speculative Inference aims to reduce repeated inference cost and has proven effective in large language models \cite{li2024eagle,li2024eagle2,li2025eagle3} and autoregressive VLAs \cite{wang2025spec,zheng2026kerv,zheng2026heisd}. Its core idea is to have a lightweight draft model propose candidates that are verified in parallel by the main model, thereby reducing expensive full generations. However, extending this paradigm to dVLAs is non-trivial.
Speculative inference requires (1) a draft proposal, (2) parallel verification, and (3) an acceptance criterion, which are all well-defined for autoregressive models through token-level probabilities. dVLAs instead produce actions through iterative denoising in continuous space \cite{chi2023diffusion,lipman2022flow}, which challenges each ingredient: it is unclear what constitutes a meaningful draft through multi-step denoising, the sequential nature of denoising resists parallel verification, and the absence of explicit likelihoods leaves no natural acceptance criterion. The central bottleneck is therefore verification: \textit{without a way to cheaply check a drafted continuous-action chunk, speculative inference collapses back to full sequential denoising.}

A key observation is that flow matching provides exactly such a structure. During flow-matching training, interpolation timesteps are sampled along linear paths between Gaussian noise and target actions, learning local flow constraints at intermediate states. Given a drafted action chunk, we can interpolate it with Gaussian noise, evaluate a few intermediate states in parallel using the Action Expert, and check whether the reconstructed endpoints agree with the draft, providing a cheap consistency check without full sequential denoising. In addition, consistency verification does not make speculative inference uniformly safe throughout the trajectory. For much of task execution, the robot performs smooth motions, where observations change slowly and small draft errors can often be tolerated. In fine-adjustment phases such as gripper switching, these errors can quickly accumulate and be amplified, leading to task failure. This motivates falling back to the full inference pipeline when verification or the task phase indicates that higher-fidelity actions are required.

In this paper, we introduce \textbf{Realtime-VLA FLASH}, a speculative inference framework for dVLAs.
FLASH introduces a fast speculative path that uses a lightweight draft model to quickly propose a candidate action chunk, and then verifies its consistency with the main model's Action Expert in parallel at a few selected timesteps, producing a dynamically sized executable action prefix.
If no action prefix is accepted, FLASH falls back to the full path to correct the trajectory.
FLASH also uses gripper switches as heuristic signals of upcoming fine-adjustment phases, where higher-quality full-path actions help prevent draft errors from accumulating into task failures.
Thus, FLASH reduces average inference latency and improves action throughput while largely preserving task success.

We evaluate Realtime-VLA FLASH in simulation and real-world tasks. 
On LIBERO, with Triton optimization, FLASH runs speculative rounds as fast as \textbf{7.8\,ms}, compared with 58.0\,ms for full-inference rounds, lowering the task-level average latency to 19.1\,ms (\textbf{3.04$\times$} speedup) and improving action throughput by \textbf{2.63$\times$}, while largely preserving task performance.
On real conveyor-belt sorting, FLASH enables successful grasping at belt speeds up to 15\,m/min, where other methods fail, demonstrating that speculative inference reduces effective inference latency and enables higher-speed, latency-sensitive manipulation tasks.

In summary, we make three contributions. First, we formulate speculative inference for flow-matching dVLAs and address its key challenge, continuous-action verification, using flow-matching interpolation paths. Second, we introduce Realtime-VLA FLASH, a dual-path inference runtime that combines a lightweight draft model, flow-consistency-based parallel verification, and phase-aware fallback to reduce expensive full-path inference during replanning. Third, we demonstrate on LIBERO and real-world tasks that FLASH replaces many expensive full-path rounds with low-cost speculative rounds, substantially reducing latency while largely preserving task success.

\section{Related Work}

\subsection{Efficient Diffusion-based VLAs}
Existing work has explored several ways to improve the efficiency of dVLAs \cite{black2024pi_0, intelligence2025pi05visionlanguageactionmodelopenworld, gr00tn1_2025,gear2025gr00tn16}, among which the following two directions are most closely related to our setting.
\paragraph{Inference Pipeline Acceleration.}
Existing work on efficient dVLAs primarily reduces the cost of the original full inference pipeline through smaller models \cite{wen2025tinyvla,shukor2025smolvla,lin2025evo}, layer compression \cite{yang2025efficientvla}, token pruning \cite{wang2025specprune}, quantization \cite{wen2025tinyvla,williams2025lite,zhang2026quantvla}, and kernel/system-level optimization \cite{ma2025running, yang2026realtime}.
These approaches improve the efficiency of the original inference pipeline, either by reducing the cost of particular components or by improving end-to-end execution efficiency.
Our work is complementary to this line of research, but it addresses a different question at the control-loop level: whether every replanning round needs to invoke the full path.

\paragraph{Diffusion and Flow-Matching Acceleration.}
Another line of work accelerates diffusion- or flow-based policies~\cite{chi2023diffusion, lipman2022flow} by shortening the generation process itself, for example through distillation into few-step models \cite{liu2026rdt2,wang2024one,frans2024one} or direct one-step model training \cite{geng2025mean, chen2026mean}. These approaches reduce the cost of sequential denoising more directly, but they typically require additional training or modified policy formulations. In contrast, we keep the original flow-matching formulation intact and improve efficiency by exploiting otherwise idle compute during the Action Denoise stage.

\subsection{Speculative Inference}

Speculative Inference accelerates autoregressive generation by letting a lightweight draft model propose candidate outputs and a larger model verify them in parallel, thereby reducing expensive decoding steps. It has been widely studied in large language models \cite{li2024eagle,li2024eagle2,li2025eagle3} and extended to autoregressive VLAs \cite{wang2025spec,zheng2026kerv,zheng2026heisd}, where discrete action tokens admit token-level probabilities that naturally support verification and acceptance.

Related acceleration has also been explored for stochastic diffusion models, where drafted denoising states can be checked using the transition structure of the stochastic reverse process~\cite{de2025accelerated,hu2025diffusion}. However, $\pi_0$ \cite{black2024pi_0}-style dVLAs differ from both settings: they generate continuous action chunks under flow matching by integrating an ODE-defined velocity field, not discrete tokens or samples from a stochastic reverse kernel. Consequently, neither token-level verification nor transition-based diffusion verification applies directly, and the absence of explicit likelihoods leaves no natural acceptance criterion for drafted continuous-action chunks.

\section{Methods}

In this section, we present the architecture of Realtime-VLA FLASH. Section~\ref{sec:method-prelim} revisits the dVLA inference pipeline and the latency bottlenecks revealed by profiling; Section~\ref{sec:method-overview} then summarizes the overall two-path design, and Sections~\ref{sec:method-draft}--\ref{sec:method-fallback} present the three core components of the flash path, namely draft action generation, multi-step parallel verification, and phase-aware fallback.

\subsection{dVLA Preliminaries and Latency Analysis}
\label{sec:method-prelim}

dVLAs, such as $\pi_0$ \cite{black2024pi_0}, typically combine a pre-trained vision-language model (VLM) backbone \cite{team2024gemma,team2024gemma2} with an Action Expert (AE). Perceptual context is passed from the VLM to the AE through the KV Cache. Formally, the model takes a multimodal observation
\[
o_t = [I_t^1,\dots,I_t^n,\ell_t,q_t],
\]
where $I_t^i$ denotes the images, $\ell_t$ the language prompt, and $q_t$ the robot state, and predicts a future action chunk
\[
A_t = [a_t, a_{t+1}, \dots, a_{t+H-1}].
\]
Each per-step action \(a_t\) contains semantically distinct channels corresponding to end-effector position \(\Delta p_t\), rotation \(\Delta \theta_t\), and binary gripper control \(g_t\). Under chunked execution, only a prefix of \(A_t\) is executed before the next action chunk is regenerated from the updated observation; we refer to each such refresh as a replanning round.

At inference time, the input images are first encoded into visual features, then concatenated with the language features, and the resulting multimodal prefix is processed by the VLM to construct a prefix KV Cache. This cache is then passed to the Action Expert, which predicts the action chunk by solving an observation-conditioned ODE from Gaussian noise:
\[
\frac{d A_t^\tau}{d\tau} = v_\theta(A_t^\tau,\tau,o_t), \qquad A_t^0 \sim \mathcal{N}(0,I),
\]
where $A_t^\tau$ denotes the intermediate noisy action state at denoising time $\tau$. At each denoising step, the Action Expert predicts the instantaneous denoising velocity field $v_\theta(A_t^\tau,\tau,o_t)$, and the final action chunk is obtained by numerically integrating this learned field over multiple timesteps, typically 10 denoising steps in the default setting, until $\tau=1$.
Consequently, a full inference round in $\pi_0$-style dVLAs consists of three stages: Image Encoder, VLM prefill to construct the visual KV Cache, and a multi-step Action Denoise stage under the learned flow field.

\begin{wrapfigure}{r}{0.45\linewidth}
    \vspace{-0.9\baselineskip}
    \centering
    \includegraphics[width=\linewidth]{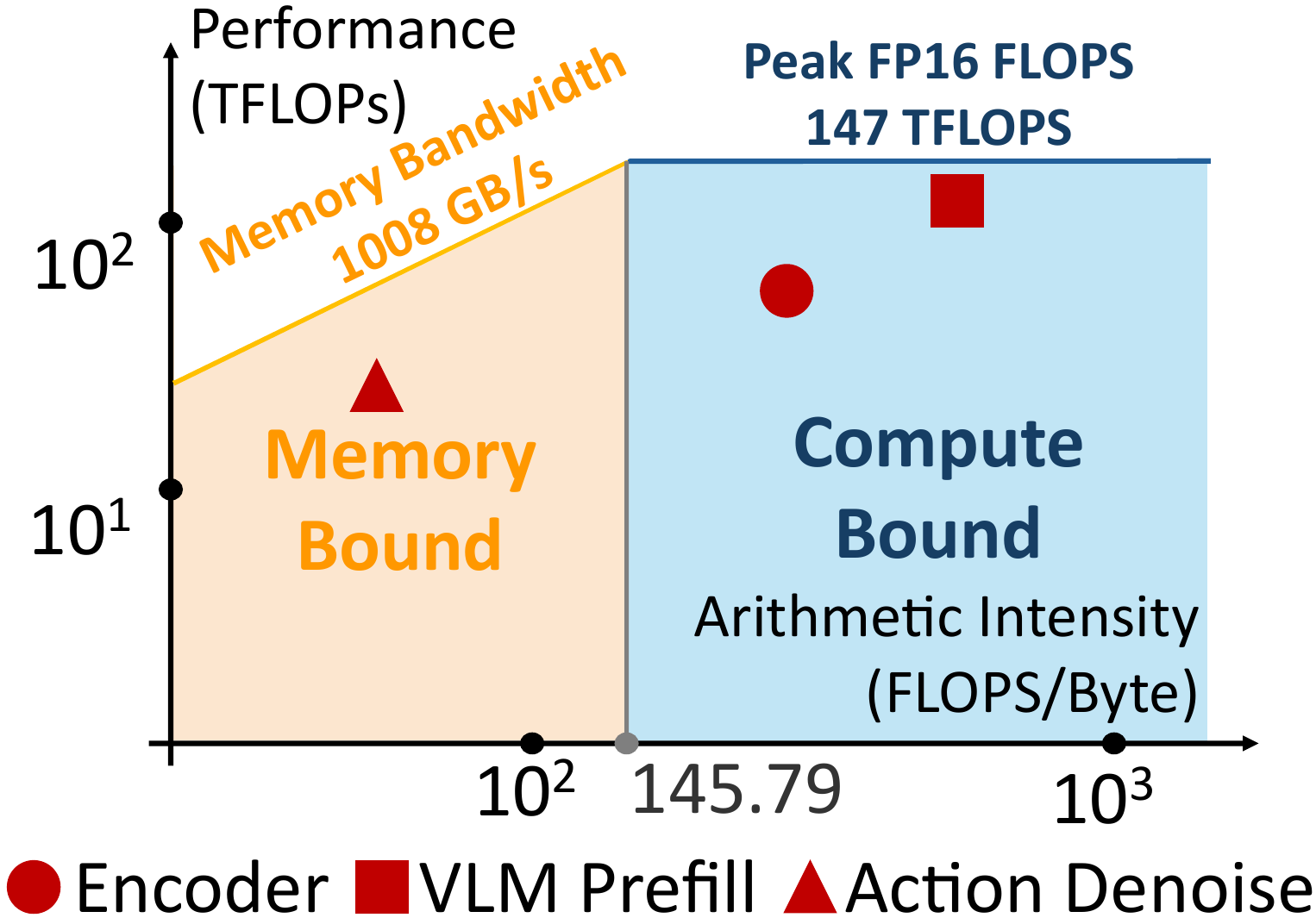}
    \caption{\textbf{Roofline analysis} of $\pi_0$ \cite{black2024pi_0} inference on NVIDIA RTX 4090D. 145.79 is the balanced arithmetic intensity point.}
    \label{fig:roofline}
    \vspace{-1.1\baselineskip}
\end{wrapfigure}

We further profile these three stages on an NVIDIA RTX 4090D \cite{jiang2026fast} and summarize the results with a roofline analysis in Figure~\ref{fig:roofline}. The measured latencies of Image Encoder, VLM prefill, and Action Denoise are 11.3\,ms, 26.7\,ms, and 20.0\,ms, respectively. The roofline places Image Encoder and VLM prefill primarily in the compute-bound regime, suggesting that their latency should be reduced by eliminating redundant computation. For VLM prefill, this naturally motivates reusing the previously computed prefix KV Cache. In contrast, Action Denoise is memory-bound, since each denoising step repeatedly reads the cache while remaining unable to parallelize across steps. This leaves available compute resources underutilized and motivates parallel verification instead of repeatedly invoking the full path.

\subsection{Realtime-VLA FLASH Overview}
\label{sec:method-overview}

\begin{figure}[!b]
    \centering
    \includegraphics[width=\linewidth]{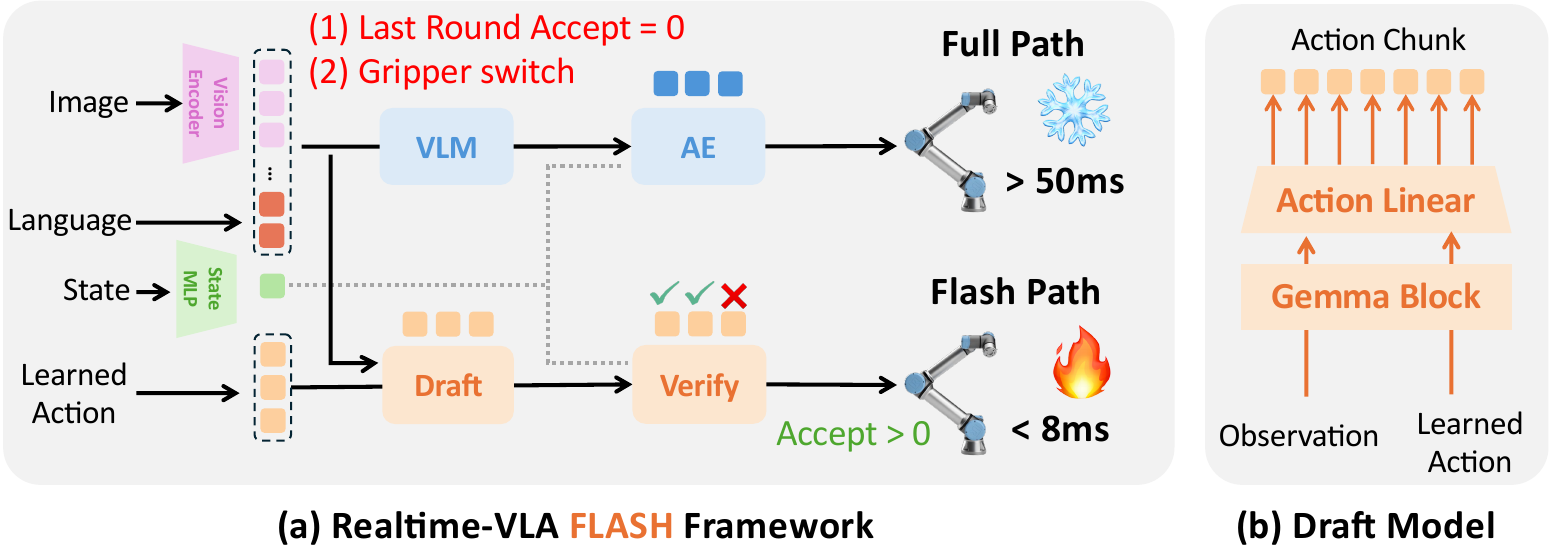}
    \caption{\textbf{Realtime-VLA FLASH Framework.} (a) Realtime-VLA FLASH uses two inference paths: the original full path (Full Path) and a lightweight speculative path (Flash Path). The Full Path performs Image Encoder, VLM prefill, and Action Denoise, while the flash path still encodes the current images but skips VLM prefill to draft a candidate action chunk for verification. (b) Draft model architecture. The draft model uses a single Gemma block with the same structure as the VLM, augments the input with learned action queries, and predicts the full action chunk in parallel through a linear action head.}
    \vskip -0.1in
    \label{fig:framework}
\end{figure}

Motivated by Section~\ref{sec:method-prelim}, Realtime-VLA FLASH adopts the dual-path architecture shown in Figure~\ref{fig:framework}(a). In a flash-path round, FLASH still encodes the current images and feeds the features to a lightweight draft model, which proposes a candidate action chunk. The main model's Action Expert then reuses the visual KV Cache produced by the most recent full-path round to reconstruct action endpoints at selected denoising timesteps and check their consistency. If no drafted action prefix is accepted, the cached context may no longer provide a reliable basis for speculative execution; FLASH therefore returns to the full path to refresh the context and correct the trajectory. In addition, FLASH uses gripper switches as phase-transition signals and proactively falls back to the full path during fine-adjustment phases, where higher-fidelity actions help mitigate error amplification.

\subsection{Draft Action Generation}
\label{sec:method-draft}

For flash-path rounds, candidate generation must be cheaper than invoking the full path. We therefore design a lightweight draft model that proposes a candidate action chunk for subsequent verification.

The draft model architecture in Figure~\ref{fig:framework}(b) consists of a single Gemma block followed by a linear action head. At inference time, the flash path still runs the image encoder of the main model on the current images and combines the resulting visual features with the language instruction, while the robot state is projected into the same hidden space through a linear layer. We then append \(H\) learned action queries to the input sequence, one for each future timestep in the chunk. The attention mask follows the blockwise structure of the base model: visual and language tokens form the conditional prefix block, the state token forms a separate block, and the action queries form the action block. The action block attends to the full conditional prefix, allowing the model to predict the entire candidate chunk in parallel. The draft model is trained by supervised regression to the target action chunk,
\[
\mathcal{L}_{\mathrm{draft}}=\sum_{h=1}^{H} w_h\,\ell\!\left(\hat a_{t+h-1}^d,\; a_{t+h-1}\right),
\]
where \(w_h\) is a step-dependent weight and \(\ell(\cdot,\cdot)\) is the per-step action regression loss.

This design has several advantages. The draft model is lightweight, containing about 110M parameters in our implementation, compared with roughly 2.7B parameters in the VLM, which makes candidate generation substantially cheaper than invoking the full path. It also remains architecturally close to the main VLM and preserves the same action-chunk representation, making it easy to integrate and allowing the verification stage to reuse the Action Expert. Additional architectural and training details are provided in Appendix~\ref{app:draft_details}.

\subsection{Multi-step Parallel Verification}
\label{sec:method-verify}
\begin{figure}[!b]
    \centering
    \includegraphics[width=\linewidth]{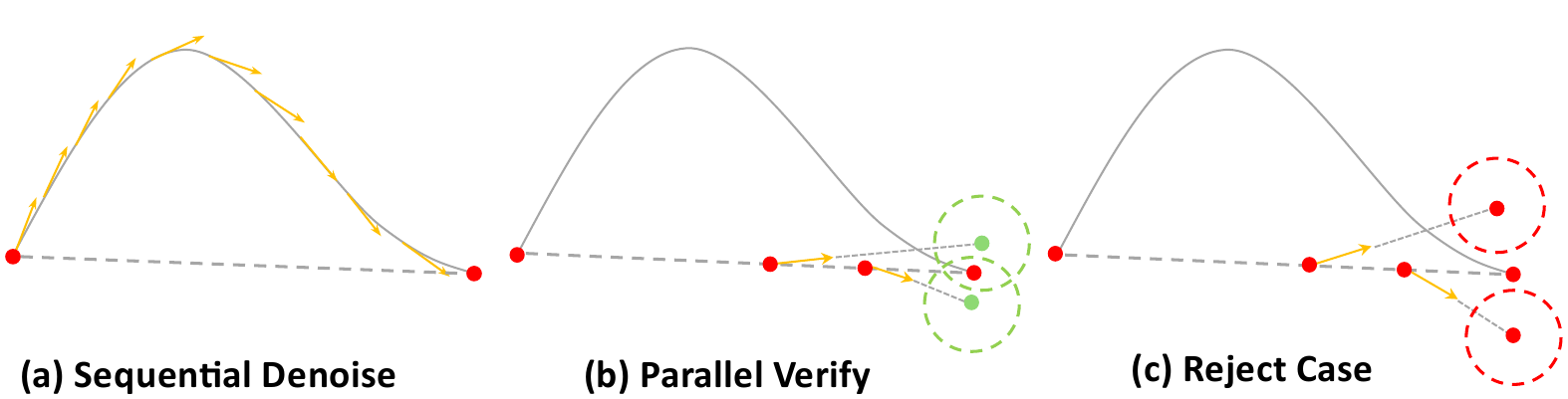}
    \caption{\textbf{Parallel verification.} (a) $\pi_0$ \cite{black2024pi_0} generates an action chunk with flow matching through 10-step sequential denoising. (b) Realtime-VLA FLASH reconstructs endpoints from selected denoising timesteps in parallel and checks an action-by-action distance threshold within the chunk, yielding the accepted prefix. (c) If reconstructed endpoints deviate beyond the threshold, no prefix is accepted and the flash path falls back to the full path.}
    \vskip -0.1in
    \label{fig:verify}
\end{figure}

The remaining challenge is to verify a draft action chunk without rerunning the full sequential denoising process in Figure~\ref{fig:verify}(a). In flow matching, this becomes possible because the policy learns the velocity field along the linear path between Gaussian noise and the endpoint at training time.

For a draft endpoint \(\hat A_t^{(d)}\) and Gaussian noise \(\epsilon\), we construct
an intermediate denoising state as
\[
\tilde A_\tau = \tau \hat A_t^{(d)} + (1-\tau)\epsilon,
\]
where \(\tau\) denotes denoising progress from Gaussian noise to the action endpoint.
Under this convention, the Action Expert predicts the local velocity from the
intermediate state toward the endpoint. Since the remaining denoising interval is \(1-\tau\), we reconstruct the endpoint by
\[
\hat A_t(\tau)
=
\tilde A_\tau
+
(1-\tau)\,
v_\theta(\tilde A_\tau,\tau \mid c_t,\, s_t).
\]
If the draft is consistent with the main policy under the reused KV Cache \(c_t\) and the latest robot state \(s_t\), these reconstructed endpoints should remain close to the draft across verification timesteps.

We therefore evaluate a small set of verification timesteps \(\mathcal{T}=\{\tau_1,\dots,\tau_K\}\) in parallel and define the executable prefix as the longest leading segment whose included actions all satisfy the distance threshold at every verification timestep, as summarized in Algorithm~\ref{alg:parallel_verify} and Figure~\ref{fig:verify}(b). We apply this distance-based rule to the continuous position and rotation channels through \(\mathrm{Dist}_{\mathrm{cont}}\), while the gripper channel is treated separately because it encodes discrete semantic states rather than incremental motion.
If \(L=0\), no executable prefix survives all verification checks, so the draft is rejected and control immediately returns to the full path, corresponding to the rejection case in Figure~\ref{fig:verify}(c).
Compared with binary acceptance, this rule returns an executable prefix, enabling more fine-grained execution and a smoother handoff to fallback. See Appendix~\ref{app:verify_consistency} for further discussion.

\begin{algorithm}[!t]
    \caption{Multi-Step Parallel Verification}
    \label{alg:parallel_verify}
    \begin{algorithmic}[1]
    \REQUIRE Draft action chunk \(\hat A_t^{(d)}\), reused visual KV Cache \(c_t\), latest robot state \(s_t\), verification timesteps \(\mathcal{T}=\{\tau_1,\dots,\tau_K\}\), threshold \(\delta\)
    \ENSURE Executable prefix length \(L\)
    \STATE Sample a shared Gaussian noise \(\epsilon\) for all verification timesteps

    \FORALL{\(\tau_k \in \mathcal{T}\) \textbf{in parallel}}
        \STATE Construct intermediate state:
        $
        \tilde A_{\tau_k}^{(k)} = \tau_k \hat A_t^{(d)} + (1-\tau_k)\epsilon
        $

        \STATE Predict local denoising velocity with the Action Expert:
        $
        v^{(k)} =
            v_\theta\!\left(\tilde A_{\tau_k}^{(k)}, \tau_k
            \,\middle|\, c_t,\, s_t\right),
        $
        \STATE Reconstruct endpoint estimate:
        $
        \hat A_t^{(k)} = \tilde A_{\tau_k}^{(k)} + (1-\tau_k)\, v^{(k)}
        $

        \FOR{\(h = 1,\dots,H\)}
            \STATE Compute step distance:
            $d_h^{(k)} = \mathrm{Dist}_{\mathrm{cont}}\!\left(\hat a_{t+h-1}^{(d)}, \hat a_{t+h-1}^{(k)}\right)$
        \ENDFOR

        \STATE Compute prefix length:
        $
        L^{(k)} = \sum_{h=1}^{H} \prod_{j=1}^{h} \mathbf{1}\!\left[d_j^{(k)} \le \delta\right]
        $
    \ENDFOR

    \STATE Return the conservative prefix length:
    $
    L = \min_{k=1,\dots,K} L^{(k)}
    $
    \end{algorithmic}
\end{algorithm}

\subsection{Phase-aware Fallback}
\label{sec:method-fallback}

\begin{figure}[!b]
    \centering
    \includegraphics[width=\linewidth]{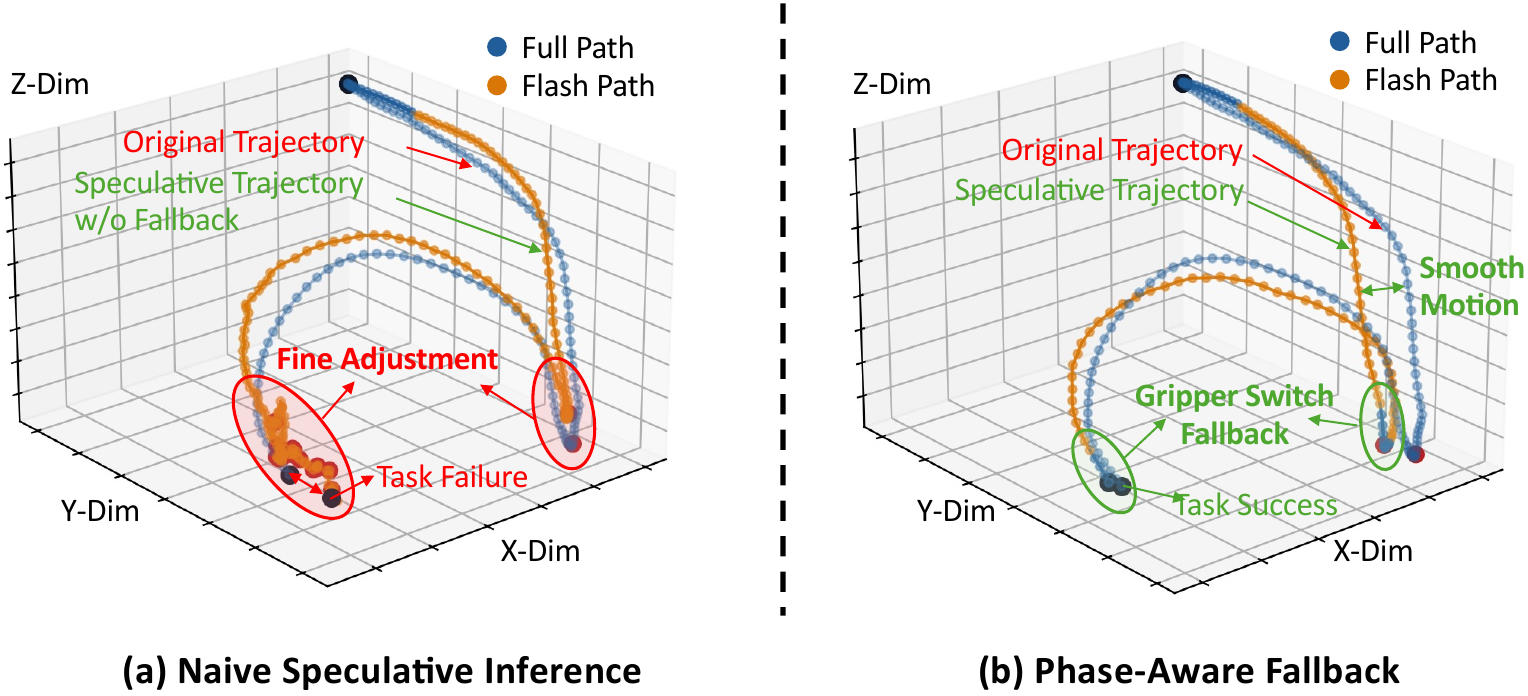}
    \caption{\textbf{Phase-aware fallback on a LIBERO-Spatial task trajectory.} 3D trajectories for a bowl-to-plate task. (a) Without phase-aware fallback, the flash path drifts during final placement and fails near the plate edge. (b) With fallback, final placement returns to the full path and succeeds.}
    \label{fig:trajectory}
\end{figure}

\begin{figure}[!t]
    \centering
    \includegraphics[width=\linewidth]{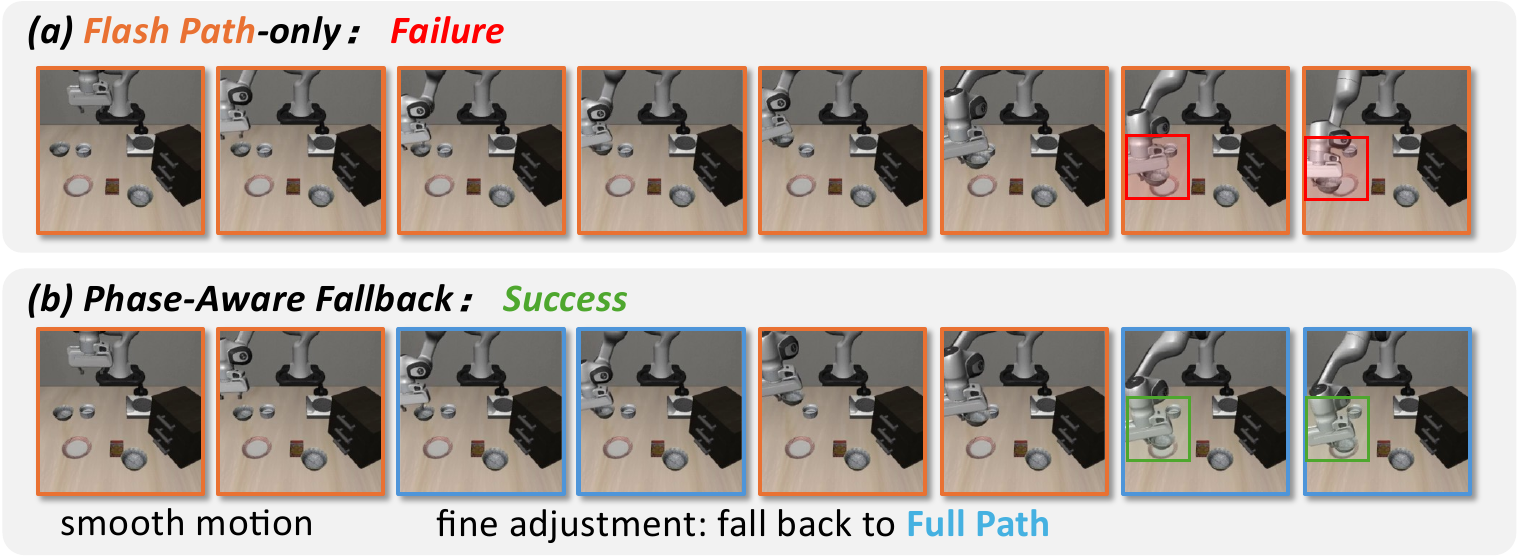}
    \caption{\textbf{Key-frame visualization of Figure~\ref{fig:trajectory}.} Orange borders indicate flash-path execution and blue borders indicate full-path execution after fallback. The failure rollout keeps final placement on the flash path, leaving the bowl misaligned with the plate.}
    \label{fig:action_frames}
  \end{figure}

Verification in Section~\ref{sec:method-verify} is local: it checks the current draft under the reused context but does not anticipate transitions into fine-adjustment phases. Figure~\ref{fig:trajectory} compares the LIBERO-Spatial bowl-to-plate trajectory without and with phase-aware fallback. Without fallback, flash-path execution continues into the final fine-adjustment phase, where the trajectory drifts toward the plate edge and leaves the bowl misaligned with the plate. Figure~\ref{fig:action_frames} provides key frames.

We therefore augment verification-triggered fallback with a phase-aware fallback rule based on the gripper channel, as shown in Figure~\ref{fig:trajectory}(b). 
Unlike the continuous position and rotation channels, the gripper channel represents discrete open and closed modes, encoded as \(-1\) and \(1\) in LIBERO. 
After applying the same standardization used by the policy, the two modes remain separated around zero, so we detect gripper switches by thresholding the standardized gripper value at zero. 
Such switches correspond to grasp or release events, which often mark precision-critical fine-adjustment phases.

If a gripper switch appears in any verification branch within the candidate
chunk, we treat it as evidence that the rollout has entered a fine-adjustment
phase. The system then falls back immediately to the full path to regenerate
higher-quality actions, avoiding error accumulation in precision-sensitive
phases where small deviations can cause task failure.

\section{Experiments}

\subsection{Setup}
\label{sec:setup}

We evaluate Realtime-VLA FLASH on $\pi_0$ \cite{black2024pi_0} in both simulation and real-world reactive manipulation tasks. All online inference experiments are run on a single NVIDIA RTX 4090D. We use an action chunk size of $H=50$ and replan after every 12 executed actions.

\textbf{Simulation.}
We follow the standard LIBERO setup \cite{liu2023libero} and report results on four suites: LIBERO-Spatial, LIBERO-Object, LIBERO-Goal, and LIBERO-10. Each suite contains 10 tasks, and each task is evaluated over 50 trials. We train a separate draft model for each LIBERO suite.

\textbf{Real-world.}
For real-world experiments, we use a single-arm UR5 platform equipped with two Intel RealSense D435i cameras, which provide multi-view observations of the workspace. We evaluate conveyor-belt sorting with two object categories: toy dog and hairbrush. For each object category, we collect 200 demonstrations and fine-tune the policy with LoRA for 20k steps. Each real-world condition is evaluated over 10 trials.
\subsection{Simulation Evaluation}

\paragraph{Setup.}
We evaluate Realtime-VLA FLASH on LIBERO by comparing four methods: Torch-$\pi_0$ \cite{black2024pi_0}, Triton-$\pi_0$ \cite{ma2025running}, FLASH-$\pi_0$, and FLASH+Triton-$\pi_0$. We adapt Triton-$\pi_0$ to LIBERO and build FLASH+Triton-$\pi_0$ by combining our FLASH with the Triton implementation. The structurally aligned draft and verify modules allow partial reuse of existing Triton-$\pi_0$ kernels. We report suite-level success rates, average task-level inference latency (\textbf{Lat.}), and per-action latency (\textbf{/Act}).

\definecolor{bestgreen}{RGB}{223,237,214}
\newcommand{\best}[1]{\textbf{#1}}
\newcommand{\pos}[1]{\textcolor{green!50!black}{#1}}
\newcommand{\negv}[1]{\textcolor{red!75!black}{#1}}

\begin{table}[tb]
    \centering
    \caption{\textbf{Simulation results on four LIBERO suites.} We report suite-level success rates, average success rate (\textbf{SR}), average task-level latency (\textbf{Lat. (ms)}), average per-action latency (\textbf{/Act (ms)}), and relative improvement over the original $\pi_0$ \cite{black2024pi_0}.}
    \label{tab:sim_main}
    \begin{tabularx}{\columnwidth}{@{}l
        *{8}{>{\hsize=.93\hsize\centering\arraybackslash}X}
        >{\hsize=1.56\hsize\centering\arraybackslash}X
        @{}}
        \toprule
        \multicolumn{1}{c}{\multirow{2}{*}[-0.15em]{\textbf{Methods}}} &
        \multicolumn{4}{c}{\textbf{Success Rate (\%)}} &
        \multicolumn{3}{c}{\textbf{Average}} &
        \multicolumn{2}{c}{\textbf{Improvement}} \\
        \cmidrule(lr){2-5}
        \cmidrule(lr){6-8}
        \cmidrule(l){9-10}
        & \textbf{Spatial} & \textbf{Object} & \textbf{Goal} & \textbf{10}
        & \textbf{SR} & \textbf{Lat.} & \textbf{/Act}
        & \textbf{$\Delta$SR} & \textbf{Speedup} \\
        \midrule
    Torch-$\pi_0$ \cite{black2024pi_0}
    & 96.8 & 98.8 & \best{95.8} & 85.2
    & 94.1 & 58.0 & 5.0
    & -- & 1.00$\times$ \\

    Triton-$\pi_0$ \cite{ma2025running}
    & 96.4 & 98.8 & 95.0 & \best{86.6}
    & 94.2 & 39.7 & 3.5
    & \pos{+0.1} & \pos{\textbf{1.46$\times$}} \\

    FLASH-$\pi_0$
    & 96.4 & 99.2 & 94.6 & 83.4
    & 93.4 & 34.9 & 3.0
    & \negv{-0.7} & \pos{\textbf{1.66$\times$}} \\

    \rowcolor{gray!12}%
    \textbf{FLASH+Triton-$\pi_0$}
    & \best{96.8} & \best{99.2} & 94.4 & 84.6
    & 93.8 & 19.1 & 1.9
    & \negv{-0.3} & \pos{\textbf{3.04$\times$}} \\
    \bottomrule
    \end{tabularx}
    \vspace{-1.0em}
\end{table}

\begin{table}[tb]
    \centering
    \caption{\textbf{Suite-level flash-path statistics on LIBERO.} \textbf{Acc.} is the average accepted draft prefix length in flash-path rounds, normalized by the replan size (12). \textbf{Flash Path Rate (FR.)} is the fraction of flash-path rounds among all replanning rounds across task trajectories.}
    \label{tab:flash_stats}
    \begin{tabular*}{\columnwidth}{@{\extracolsep{\fill}}lcccccccc}
        \toprule
        \multicolumn{1}{c}{\multirow{2}{*}[-0.15em]{\textbf{LIBERO Suites}}} &
        \multicolumn{2}{c}{\textbf{Acc. (\%)}} &
        \multicolumn{2}{c}{\textbf{FR. (\%)}} &
        \multicolumn{2}{c}{\textbf{Lat. (ms)}} &
        \multicolumn{2}{c}{\textbf{/Act (ms)}} \\
        \cmidrule(lr){2-3}
        \cmidrule(lr){4-5}
        \cmidrule(lr){6-7}
        \cmidrule(l){8-9}
        & \textbf{FLASH} & \textbf{+Tri.}
        & \textbf{FLASH} & \textbf{+Tri.}
        & \textbf{FLASH} & \textbf{+Tri.}
        & \textbf{FLASH} & \textbf{+Tri.} \\
        \midrule
        Spatial & 75.8 & 60.2 & 71.6 & 68.9 & 34.3 & 19.2 & 3.0 & 2.1 \\
        Object & 78.1 & \best{89.3} & \best{72.4} & \best{86.4} & 34.6 & \best{12.6} & 3.0 & \best{1.2} \\
        Goal & 81.4 & 66.9 & 63.7 & 62.4 & 37.5 & 20.5 & 3.3 & 2.1 \\
        10 & \best{82.4} & 62.5 & 57.5 & 49.6 & \best{33.2} & 24.1 & \best{2.8} & 2.3 \\
        \midrule
        \textbf{Avg.} & 79.4 & 69.7 & 66.3 & 66.8 & 34.9 & 19.1 & 3.0 & 1.9
        \\
        \bottomrule
    \end{tabular*}
    \vspace{-1.0em}
\end{table}

\paragraph{Results.} Table~\ref{tab:sim_main} summarizes the main simulation results.
Compared with the 58.0\,ms original full-path round, a flash-path round costs only 17.9\,ms for FLASH-\(\pi_0\) and \textbf{7.8\,ms} for FLASH+Triton-\(\pi_0\) (Appendix~\ref{app:draft_details}, Table~\ref{tab:runtime_cost}).
With phase-aware fallback, FLASH-\(\pi_0\) reduces the task-level average latency to 34.9\,ms, achieving a 1.66$\times$ speedup over Torch-\(\pi_0\); with Triton optimization, FLASH+Triton-\(\pi_0\) further reduces it to \textbf{19.1\,ms}, achieving a \textbf{3.04$\times$} speedup.
This also corresponds to a \textbf{2.63$\times$} reduction in per-action latency, with only a 0.3-point drop in average success rate.


\begin{wraptable}{r}{0.44\linewidth}
    \centering
    \scriptsize
    \captionsetup{width=\linewidth}
    \caption{\textbf{Ablation of key components on LIBERO-10.} PF = $n$ denotes forcing one full-path refresh every $n$ flash-path rounds.}
    \label{tab:components_ablation}
    \begin{tabular}{lcc}
        \toprule
        \textbf{Configuration} & \textbf{SR (\%)} & \textbf{Lat. (ms)}\\
        \midrule
        Baseline (Flash-path only) & 58.4 & \textbf{13.3} \\
        \midrule
        \textit{Periodic full-path refresh} & & \\
        + PF = 2 & 80.6 & 21.0 \\
        + PF = 3 & 82.0 & 18.6 \\
        + PF = 4 & 75.4 & 17.2 \\
        \midrule
        \textit{Phase-aware fallback} & & \\
        + FB & 66.8 & 17.7 \\
        \rowcolor{gray!12}
        \textbf{+ FB \& PF = 2} & \textbf{84.6} & 24.1 \\
        + FB \& PF = 3 & 81.6 & 21.9 \\
        + FB \& PF = 4 & 78.4 & 20.2 \\
        \bottomrule
    \end{tabular}
    \vspace{-1.0em}
\end{wraptable}

Table~\ref{tab:flash_stats} attributes the FLASH speedup to high flash-path usage: FLASH+Triton-$\pi_0$ handles 66.8\% of replanning rounds via the flash path, with accepted prefixes covering 69.7\% of the replan window, thereby reducing full-path calls and amortizing speculative execution.
Table~\ref{tab:components_ablation} evaluates runtime components on LIBERO-10, the most hyperparameter-sensitive suite, where small speculative errors can accumulate and cause task failure. We fix the verification threshold to $\delta=0.15$ and use two verifier timesteps ($|\mathcal{T}|=2$), then ablate periodic full-path refresh (\textbf{PF}) and phase-aware fallback (\textbf{FB}). PF periodically invokes higher-fidelity full-path actions to correct long-horizon drift, while FB performs targeted correction near precision-critical phases indicated by gripper switches. We finally select +FB \& PF=2 as the best success-latency trade-off for the LIBERO-10 result in Table~\ref{tab:sim_main}, with detailed ablations provided in Appendix~\ref{app:ablation}.

\subsection{Real-world Evaluation}

\begin{figure}[!t]
    \centering
    \includegraphics[width=0.8\linewidth]{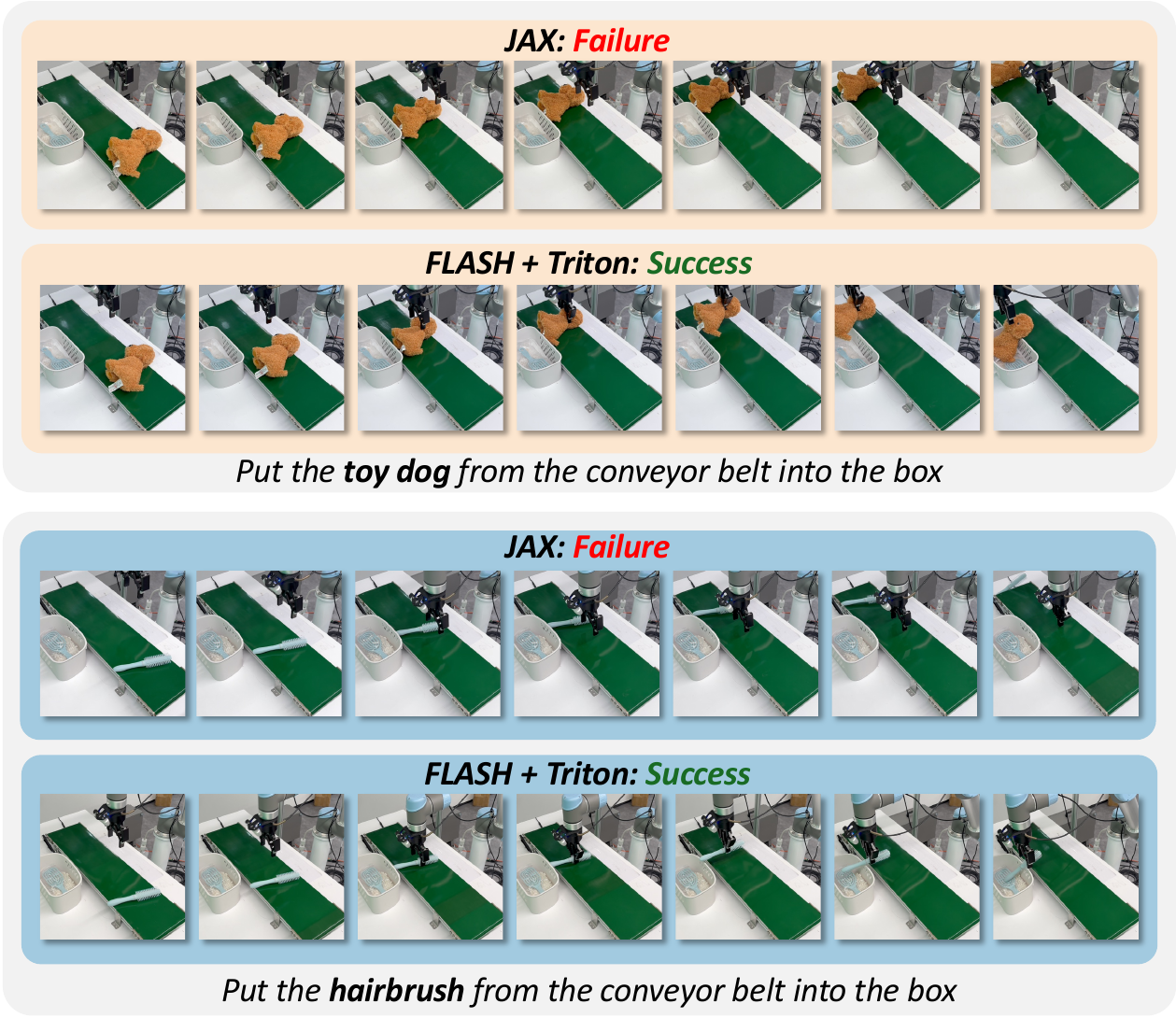}
    \caption{Conveyor-belt sorting demo.}
    \label{fig:real_demo}
    \vspace{-1.0em}
\end{figure}

\paragraph{Setup.}
We evaluate real-world conveyor-belt sorting, where the robot must grasp a moving object and finish sorting before it leaves the reachable region, as shown in Figure~\ref{fig:real_demo}. To isolate policy inference latency, all methods use the same synchronous inference loop without asynchronous inference, real-time chunking~\cite{black2025real, black2025training}, or other latency-hiding optimizations. We compare JAX-$\pi_0$ \cite{black2024pi_0}, Triton-$\pi_0$ \cite{ma2025running}, and FLASH+Triton-$\pi_0$ under the same hardware and protocol across medium, high, and extra high speeds. Appendix~\ref{app:real_details} provides the platform and training hyperparameters.

\begin{table}[!t]
    \centering
    \caption{\textbf{Conveyor-belt sorting results under synchronous inference.} We report success rate (\%) over 10 trials for each object and speed; demonstrations are collected at 6\,m/min.}
    \label{tab:real_sorting}
    \begin{tabularx}{\columnwidth}{@{}l*{6}{>{\centering\arraybackslash}X}@{}}
        \toprule
        \multicolumn{1}{c}{\multirow{2}{*}[-0.15em]{\textbf{Methods}}} &
        \multicolumn{2}{c}{\textbf{Medium (10\,m/min)}} &
        \multicolumn{2}{c}{\textbf{High (13\,m/min)}} &
        \multicolumn{2}{c}{\textbf{Extra High (15\,m/min)}} \\
        \cmidrule(lr){2-3}
        \cmidrule(lr){4-5}
        \cmidrule(l){6-7}
        & \textbf{Toy dog} & \textbf{Hairbrush}
        & \textbf{Toy dog} & \textbf{Hairbrush}
        & \textbf{Toy dog} & \textbf{Hairbrush} \\
        \midrule
        JAX-$\pi_0$ \cite{black2024pi_0} & 20.0 & 50.0 & 0.0 & 0.0 & 0.0 & 0.0 \\
        Triton-$\pi_0$ \cite{ma2025running} & 80.0 & 40.0 & 30.0 & 10.0 & 0.0 & 0.0 \\
        \rowcolor{gray!12}%
        \textbf{FLASH+Triton-$\pi_0$} & \best{80.0} & \best{90.0} & \best{50.0} & \best{30.0} & \best{20.0} & \best{10.0} \\
        \bottomrule
    \end{tabularx}
    \vspace{-1.0em}
\end{table}

\paragraph{Results.}
Table~\ref{tab:real_sorting} shows that conveyor speed amplifies the effect of policy inference latency.
As a reference point below the reported speed range, JAX-\(\pi_0\) achieves 80\% success on toy dog at 8\,m/min, but drops sharply at the reported speeds.
At medium speed, all methods achieve nonzero success, but FLASH+Triton-$\pi_0$ already improves the more timing-sensitive hairbrush task. As speed increases, the synchronous baselines degrade rapidly: JAX-$\pi_0$ \cite{black2024pi_0} fails at high speed, Triton-$\pi_0$ \cite{ma2025running} retains only limited success, and both fail at extra high speed. In contrast, FLASH+Triton-$\pi_0$ remains successful at extra high speed and is the only method with nonzero success at 15\,m/min.

Failures of the slower baselines mostly come from stale action chunks: the robot approaches an outdated belt position, so the gripper arrives behind the object or closes too late. This effect is stronger at higher speeds and on hairbrushes, whose thin geometry leaves less tolerance to timing and pose errors. These failures support the main real-world takeaway: under synchronous control, reducing policy latency directly expands the speed range in which dVLAs can complete reactive manipulation.

\section{Conclusion}
We presented Realtime-VLA FLASH, a speculative inference framework that serves most replanning rounds through the flash path and reverts to the full path when verification or fine-adjustment phases require higher fidelity. Experiments on LIBERO and conveyor-belt sorting show substantial latency reductions while largely preserving task performance, suggesting a practical path toward latency-sensitive embodied control. FLASH still relies on heuristic thresholds, leaving trajectory-adaptive verification thresholds as an important direction for future work.
\section*{Acknowledgements}
We thank Haoqiang Fan and collaborators for insightful discussions and for the Realtime-VLA line of work~\cite{ma2025running,yang2026realtime} which helped shape our focus on latency-aware VLA deployment.

\medskip

\bibliography{neurips_2026}
\bibliographystyle{plainnat}


\appendix

\section{Draft Model Details}
\label{app:draft_details}

\subsection{Architecture}

This section expands the draft architecture introduced in Section~\ref{sec:method-draft}.
Because the draft model reuses a VLM block, the observations provide visual context but do not define where a fixed-length action chunk should be written. 
We therefore append \(H\) learned action slots as explicit output positions:
\[
Q = [q_1, q_2, \dots, q_H],
\]
where \(q_h\) corresponds to the \(h\)-th future action. 
These slots are trainable parameters, and they serve as persistent output queries that decode structured future actions from the shared context.

We use a blockwise attention mask to make these slots compatible with the reused VLM architecture. 
Each action slot attends to the visual-language prefix and state token, and the slots can also interact within the action block, allowing the predicted actions to be coordinated across the chunk. 
Because all \(H\) slots are processed in the same forward pass, the transformer produces hidden states for the entire action chunk in parallel.

After the transformer block, only the hidden states associated with the action slots are decoded. 
A shared linear action head maps each slot representation to its corresponding continuous action:
\[
\hat a_{t+h-1}^{d} = W_{\mathrm{act}} z_h + b_{\mathrm{act}}, \qquad h=1,\dots,H,
\]
where \(z_h\) is the final hidden representation of slot \(q_h\).

Thus, the learned action slots convert a VLM block into a parallel action-chunk predictor. 

\subsection{Training Hyperparameters and Inference Cost Breakdown}

Table~\ref{tab:draft_training_config} summarizes the draft-model training setup.
The draft model regresses to frozen full-path policy outputs with a prefix-weighted objective, so early executable actions receive higher weight while verification and fallback still decide the final executed prefix.

\begin{table}[!h]
    \vspace{-1.0em}
    \centering
    \scriptsize
    \caption{Training hyperparameters for the draft model.}
    \label{tab:draft_training_config}
    \begin{tabularx}{\columnwidth}{@{}lX@{}}
    \toprule
    \textbf{Item} & \textbf{Value} \\
    \midrule
    Optimizer & AdamW \\
    Learning rate & \(2\times10^{-3}\) \\
    Weight decay & \(0.01\) \\
    Batch size & 64 \\
    Epochs & 100 \\
    Precision & FP32 \\
    \midrule
    Loss & Weighted Smooth L1 / Huber loss, \(\beta=1.0\) \\
    Prefix weighting & Sampled prefix up to 16 steps, \(\gamma_{\mathrm{prefix}}=0.9\), tail weight \(0.1\) \\
    Checkpoint & Best validation RMS over the first 12 executable steps \\
    \midrule
    Compute cost & 4 NVIDIA RTX 4090D GPUs, approximately 6 hours per draft model \\
    \bottomrule
    \end{tabularx}
\end{table}

Table~\ref{tab:runtime_cost} reports the component-level latency behind the task-level results in Table~\ref{tab:sim_main}. We include both per-round component costs and trajectory-level averages because flash-path rounds may execute variable-length prefixes.

\begin{table}[tb]
    \scriptsize
    \centering
    \caption{Inference cost breakdown.}
    \label{tab:runtime_cost}
    \begin{tabular}{lcccc}
        \toprule
        \textbf{Component} 
        & \textbf{Torch-\(\pi_0\)}
        & \textbf{Triton-\(\pi_0\)}
        & \textbf{FLASH-\(\pi_0\)}
        & \textbf{FLASH+Triton-\(\pi_0\)} \\
        \midrule
        \multicolumn{5}{l}{\textit{Full-path round (ms)}} \\
        Image encoder & 11.3 & 4.7 & 11.3 & 4.7 \\
        VLM prefill & 26.7 & 22.4 & 26.7 & 22.4 \\
        Action denoise & 20.0 & 12.6 & 20.0 & 12.6 \\
        Full-path total & 58.0 & 39.7 & 58.0 & 39.7 \\
        \midrule
        \multicolumn{5}{l}{\textit{Flash-path round (ms)}} \\
        Image encoder & -- & -- & 11.0 & 4.7 \\
        Draft model & -- & -- & 3.5 & 0.9 \\
        Parallel verifier & -- & -- & 3.4 & 2.2 \\
        Flash-path total & -- & -- & \textbf{17.9} & \textbf{7.8} \\
        \midrule
        \multicolumn{5}{l}{\textit{Trajectory-level average}} \\
        Replanning latency & 58.0 & 39.7 & 34.9 & 19.1 \\
        Per-action latency & 5.0 & 3.5 & 3.0 & 1.9 \\
        \rowcolor{gray!12}
        Speedup & 1.00\(\times\) & 1.46\(\times\) & 1.66\(\times\) & \textbf{3.04\(\times\)} \\
        \bottomrule
    \end{tabular}
    \vspace{-1.0em}
\end{table}

\section{Verification Consistency Interpretation}
\label{app:verify_consistency}

The endpoint check in Algorithm~\ref{alg:parallel_verify} should be interpreted
as a heuristic local consistency test, rather than as a formal guarantee that the
accepted draft and the full-path rollout induce identical trajectories or
task-level behavior. The key distinction is that the Action Expert is trained by
flow matching on interpolation paths toward target action endpoints, whereas
verification probes the learned flow field on states induced by a drafted
endpoint. Thus, passing verification indicates that the draft is locally
compatible with the learned velocity field under the reused conditioning context,
but it does not establish full-path equivalence.

Let \(A_t^\star\) denote the endpoint produced by a full-path rollout under the
current conditioning context, and let \(\hat A_t^{(d)}\) denote the accepted
draft endpoint. Informally, the endpoint discrepancy can be viewed as
controlled by the acceptance threshold, the local reconstruction error of the
Action Expert, the residual conditioning mismatch, and the mismatch between
target-induced and draft-induced interpolation paths:
\[
\|\hat A_t^{(d)} - A_t^\star\|
\lesssim
\delta
+
\epsilon_{\mathrm{AE}}
+
\epsilon_{\mathrm{cond}}
+
\epsilon_{\mathrm{path}} .
\]
Here, \(\epsilon_{\mathrm{AE}}\) accounts for approximation errors in the local
velocity field and the single-step endpoint reconstruction performed by the
Action Expert. The term \(\epsilon_{\mathrm{cond}}\) captures residual mismatch
introduced by reusing the cached visual-language prefix. The term
\(\epsilon_{\mathrm{path}}\) captures the fact that verification evaluates the
flow field on draft-induced interpolation states, rather than on the
target-endpoint paths used during training or the states visited by a full
sequential rollout.

Importantly, as shown in Algorithm~\ref{alg:parallel_verify}, the robot state
\(s_t\) is refreshed at every verification call and therefore does not constitute
a stale cached variable. Overall, the acceptance rule should be understood as a
conservative local agreement criterion, not as a formal correctness guarantee for
the accepted prefix. This interpretation motivates conservative thresholds,
multi-timestep verification, phase-aware fallback, and full-path fallback when
local consistency fails.
\section{Experimental Details}
\label{app:exp}

\subsection{Simulation Details: Verification Hyperparameter Ablation}

\label{app:ablation}

We isolate the verifier before adding phase-aware fallback and periodic full-path refresh, so this subsection should be interpreted as a verifier trade-off study.
Table~\ref{tab:verify_ablation} uses LIBERO-10 as a stress test and varies the number of verification timesteps \(K=|\mathcal{T}|\) and distance threshold \(\delta\). 
Loose thresholds keep more rounds on the flash path and reduce latency, but they also allow inaccurate drafts to pass verification. 
Stricter thresholds and larger \(K\) reject more speculative actions, improving reliability in sensitive settings while pushing latency toward the full-path regime.

We use \(K=2,\delta=0.15\) as the base verifier setting for the full runtime rather than as a standalone verifier-only optimum. 
This point preserves high flash-path coverage and low latency, while the complete runtime recovers reliability through phase-aware fallback and periodic full-path refresh (Table~\ref{tab:components_ablation}). 
Table~\ref{tab:delta_suite_ablation} further fixes \(K=2\) and sweeps \(\delta\) across all LIBERO suites. 
The same acceptance--latency pattern appears across suites, while LIBERO-10 shows the largest success-rate variation; we therefore use it as the primary verifier stress test in Table~\ref{tab:verify_ablation}.

\begin{table}[!h]
    \vspace{-1.0em}
    \centering
    \caption{Verifier-only ablation on LIBERO-10.}
    \label{tab:verify_ablation}
    \begin{threeparttable}
    \scriptsize
    \begin{tabularx}{\columnwidth}{@{}*{8}{>{\centering\arraybackslash}X}@{}}
        \toprule
        \multicolumn{3}{c}{\textbf{Verifier Setting}} &
        \multicolumn{5}{c}{\textbf{LIBERO-10 Result}} \\
        \cmidrule(lr){1-3}
        \cmidrule(l){4-8}
        \textbf{\(K\)} & \textbf{Lat. (ms)} & \(\boldsymbol{\delta}\)
        & \textbf{SR (\%)} & \textbf{Lat. (ms)} & \textbf{/Act (ms)}
        & \textbf{Acc. (\%)} & \textbf{FR. (\%)} \\
        \midrule
        \multirow{5}{*}{4} & \multirow{5}{*}{4.3}
        & 0.20 & 82.6 & 35.7 & 4.2 & 13.9 & 18.8 \\
        & & 0.15 & 86.2 & 39.9 & 4.4 & 6.8 & 9.2 \\
        & & 0.10 & 85.0 & 47.5 & 4.5 & 1.8 & 2.1 \\
        & & 0.05 & 84.4 & 52.4 & 4.5 & 0.0 & 0.0 \\
        & & 0.00 & 86.4 & 52.3 & 4.5 & 0.0 & 0.0 \\
        \midrule
        \multirow{5}{*}{2} & \multirow{5}{*}{2.2}
        & 0.20 & 54.3 & 10.3 & 0.9 & 91.8 & 94.3 \\
        & & \textbf{0.15} & \textbf{58.4} & \textbf{13.3} & \textbf{1.5} & \textbf{65.6} & \textbf{83.0} \\
        & & 0.10 & 79.4 & 21.1 & 3.3 & 23.3 & 40.9 \\
        & & 0.05 & 93.5 & 44.9 & 4.5 & 2.0 & 2.5 \\
        & & 0.00 & 85.6 & 47.9 & 4.1 & 0.0 & 0.0 \\
        \midrule
        \multirow{5}{*}{1} & \multirow{5}{*}{1.4}
        & 0.20 & 54.9 & 13.5 & 1.2 & 99.7 & 95.4 \\
        & & 0.15 & 53.8 & 12.5 & 1.1 & 99.4 & 95.4 \\
        & & 0.10 & 51.0 & 12.9 & 1.2 & 84.8 & 92.4 \\
        & & 0.05 & 84.2 & 28.5 & 4.0 & 11.4 & 19.6 \\
        & & 0.00 & 86.8 & 49.8 & 4.2 & 0.0 & 0.0 \\
        \bottomrule
    \end{tabularx}
    \end{threeparttable}
\end{table}

\begin{table}[!h]
    \vspace{-1.0em}
    \centering
    \scriptsize
    \caption{Cross-suite threshold sensitivity with $K=2$.}
    \label{tab:delta_suite_ablation}
    \begin{threeparttable}
    \resizebox{\columnwidth}{!}{%
    \begin{tabular}{@{}lrrrrrrrrrrrrr@{}}
        \toprule
        \multirow{2}{*}{\(\boldsymbol{\delta}\)} &
        \multicolumn{4}{c}{\textbf{Success Rate (\%)}} &
        \multicolumn{4}{c}{\textbf{Latency (ms)}} &
        \multicolumn{5}{c}{\textbf{Average}} \\
        \cmidrule(lr){2-5}
        \cmidrule(lr){6-9}
        \cmidrule(l){10-14}
        & \textbf{Spatial} & \textbf{Object} & \textbf{Goal} & \textbf{10}
        & \textbf{Spatial} & \textbf{Object} & \textbf{Goal} & \textbf{10}
        & \textbf{SR} & \textbf{Lat.} & \textbf{/Act} & \textbf{Acc.} & \textbf{FR.} \\
        \midrule
        0.30 & 95.8 & 99.2 & 86.6 & 53.8 & 12.0 & 11.1 & 11.8 &  9.9 & 83.9 & 11.1 & 1.0 & 99.7 & 91.3 \\
        0.25 & 93.6 & 98.4 & 87.6 & 53.8 & 11.9 & 11.3 & 11.8 & 10.0 & 83.4 & 11.1 & 1.0 & 98.8 & 91.4 \\
        0.20 & 94.8 & 98.6 & 86.4 & 54.3 & 11.8 & 11.3 & 11.9 & 10.3 & 83.5 & 11.2 & 1.0 & 95.1 & 90.9 \\
        \rowcolor{gray!12}
        \textbf{0.15} & \textbf{96.0} & \textbf{99.2} & \textbf{90.8} & \textbf{58.4} & \textbf{12.3} & \textbf{12.6} & \textbf{12.5} & \textbf{13.3} & \textbf{86.1} & \textbf{12.7} & \textbf{1.4} & \textbf{77.5} & \textbf{84.9} \\
        0.10 & 97.2 & 98.8 & 92.6 & 79.4 & 19.5 & 20.7 & 19.5 & 21.1 & 92.0 & 20.4 & 3.0 & 30.5 & 46.6 \\
        0.05 & 93.8 & 98.8 & 94.4 & 87.2 & 41.0 & 40.2 & 40.9 & 41.6 & 93.5 & 41.0 & 4.1 &  2.0 &  2.5 \\
        0.00 & 97.0 & 97.6 & 96.2 & 85.6 & 47.4 & 47.2 & 47.2 & 47.9 & 94.1 & 47.5 & 4.1 &  0.0 &  0.0 \\
        \bottomrule
    \end{tabular}}
    \end{threeparttable}
    \vspace{-1.0em}
\end{table}

\subsection{Real-world Details}
\label{app:real_details}

\begin{figure}[!h]
    \vspace{-1.0em}
    \centering
    \includegraphics[width=0.65\linewidth]{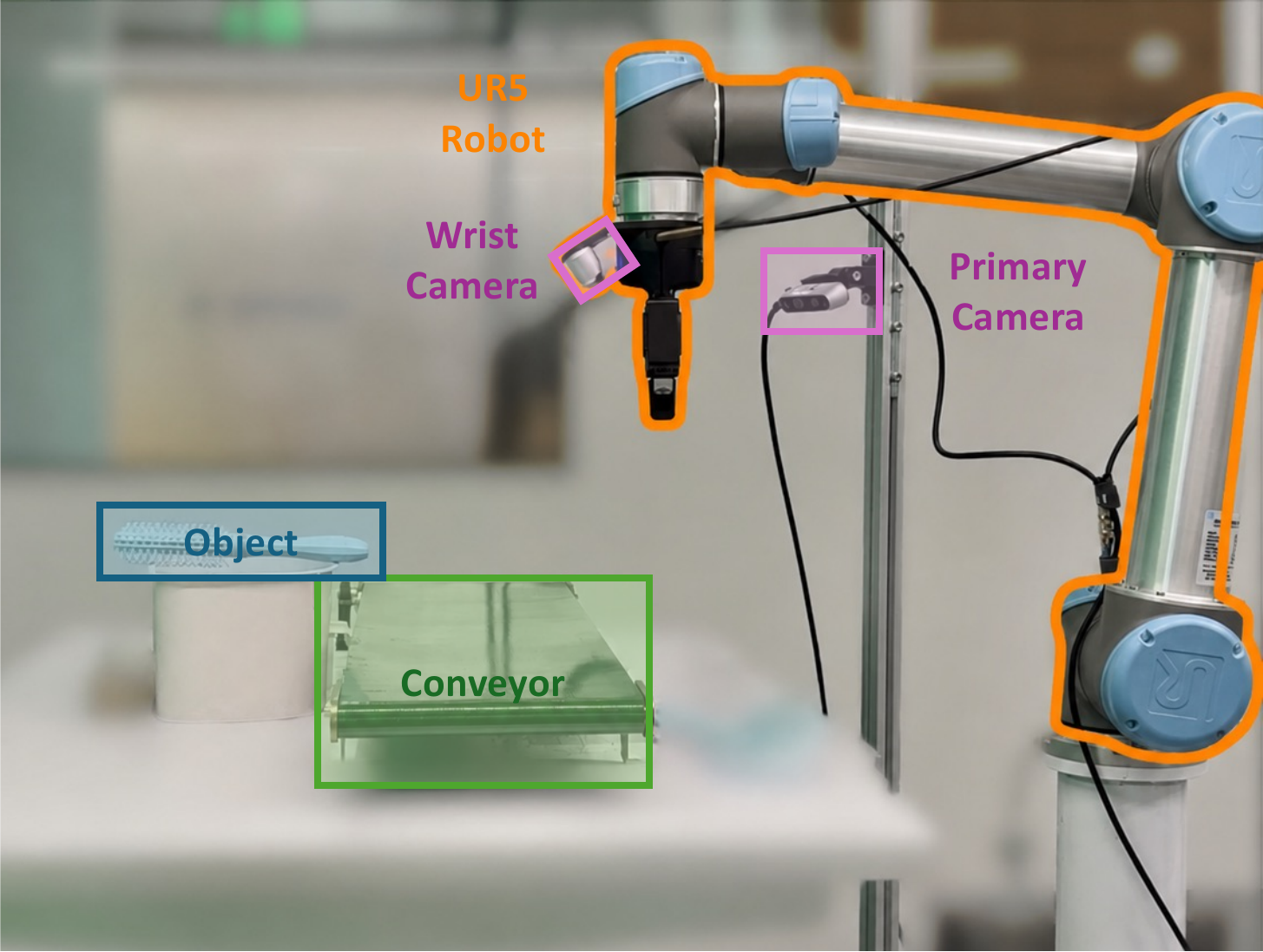}
    \caption{Real-world robot platform used for conveyor-belt sorting.}
    \label{fig:real_robot}
    \vspace{-1.0em}
\end{figure}

Figure~\ref{fig:real_robot} illustrates the real-world evaluation platform used for conveyor-belt sorting.
In the real-world deployment, accepted actions are still executed through the robot controller with joint limits and command constraints, so a false acceptance remains subject to these low-level safety constraints.

Table~\ref{tab:real_training_hparams} reports the training configuration.
We first fine-tune the main policy from the pre-trained \(\pi_0\) checkpoint using LoRA on the collected demonstrations, and then train the draft model for the same real-world domain.
For draft training, the observations come from the real demonstrations, while the regression targets are teacher action chunks generated by the fine-tuned main model rather than directly by behavior cloning from raw demonstration actions.

\begin{table}[!h]
    \centering
    \scriptsize
    \caption{Training hyperparameters for real-world demonstration adaptation.}
    \label{tab:real_training_hparams}
    \begin{tabularx}{\columnwidth}{@{}l>{\raggedright\arraybackslash}X@{}}
        \toprule
        \textbf{Item} & \textbf{Value} \\
        \midrule
        \multicolumn{2}{@{}l}{\textit{Main Model fine-tuning}} \\
        Batch size & 128 \\
        Training steps & 20k \\
        LR schedule & Cosine decay with 500 warmup steps \\
        Peak / final LR & \(1\times10^{-4}\) / \(1\times10^{-5}\) \\
        \midrule
        \multicolumn{2}{@{}l}{\textit{Draft model training}} \\
        Optimizer & AdamW \\
        Learning rate & \(2\times10^{-3}\) \\
        LR schedule & Cosine schedule with 1 warmup epoch and \(1\times10^{-5}\) minimum LR \\
        Batch size & 64 per GPU \\
        Epochs & 100 \\
        Gradient clipping & Global norm 1.0 \\
        Loss & Weighted Smooth L1 / Huber loss, \(\beta=1.0\) \\
        Prefix weighting & Sampled prefix up to 20 steps, \(\gamma_{\mathrm{prefix}}=0.9\), tail weight \(0.1\) \\
        Checkpoint & Best validation RMS over the first 12 executable steps \\
        \midrule
        Compute cost & 8 NVIDIA H20 GPUs, approximately 12 hours for fine-tuning and 2 hours for draft model training \\
        \bottomrule
    \end{tabularx}
\end{table}

\section{Future Work}
\label{app:future_work}

Realtime-VLA FLASH suggests several directions for future work. 

First, the current verifier uses heuristic hyperparameters, including fixed acceptance thresholds and hand-picked verification timesteps. A more adaptive verifier could adjust both the threshold and the timestep locations over the course of a trajectory, tightening verification during fine-adjustment phases, rapid observation changes, or high-curvature motions, while relaxing it during smooth free-space motion. This generalizes the intuition behind our phase-aware fallback mechanism from a gripper-switch heuristic to a broader context-dependent verification rule.

Second, FLASH could be combined with real-time chunking (RTC) methods. Current RTC methods often assume a fixed inference latency, whereas FLASH introduces dynamic accepted prefix lengths and variable effective latency across replanning rounds. Adapting RTC to this setting would require trajectory optimization algorithms that account for variable-latency policy updates, potentially yielding trajectories that are both fast and smooth.

Third, FLASH may further benefit edge deployments. VLA-Perf~\cite{jiang2026fast} reports that VLA inference stages become memory-bound on edge devices, where memory bandwidth, power, and thermal budgets are substantially tighter than on desktop GPUs. By reducing repeated full-path inference calls, FLASH can lower average memory traffic and power consumption during replanning. Extending this design to resource-constrained edge devices could make low-latency dVLA deployment more practical for robots with limited compute, power, or thermal budgets.



\end{document}